\documentclass{article} 
\usepackage{colm2024_conference}

\usepackage{microtype}
\usepackage{hyperref}
\usepackage{url}
\usepackage{booktabs}
\usepackage{amsmath}
\usepackage{graphicx}
\usepackage{makecell}
\usepackage{subcaption}

\title{Uncertainty Estimation of Large Language Models in Medical Question Answering}

\author{Jiaxin Wu \\
The University of Hong Kong\\
\texttt{lisa24@connect.hku.hk} 
\AND
Yizhou Yu \\
The University of Hong Kong\\
\texttt{yizhouy@acm.org} 
\AND
Hong-Yu Zhou \\
The University of Hong Kong \& Harvard Medical School\\
\texttt{whuzhouhongyu@gmail.com} 
}

\colmfinalcopy 
\begin{document}

\maketitle

\begin{abstract}
Large Language Models (LLMs) show promise for natural language generation in healthcare, but risk hallucinating factually incorrect information. Deploying LLMs for medical question answering necessitates reliable uncertainty estimation (UE) methods to detect hallucinations. In this work, we benchmark popular UE methods with different model sizes on medical question-answering datasets. Our results show that current approaches generally perform poorly in this domain, highlighting the challenge of UE for medical applications. We also observe that larger models tend to yield better results, suggesting a correlation between model size and the reliability of UE. To address these challenges, we propose Two-phase Verification, a probability-free Uncertainty Estimation approach. First, an LLM generates a step-by-step explanation alongside its initial answer, followed by formulating verification questions to check the factual claims in the explanation. The model then answers these questions twice: first independently, and then referencing the explanation. Inconsistencies between the two sets of answers measure the uncertainty in the original response. We evaluate our approach on three biomedical question-answering datasets using Llama 2 Chat models and compare it against the benchmarked baseline methods. The results show that our Two-phase Verification method achieves the best overall accuracy and stability across various datasets and model sizes, and its performance scales as the model size increases.
\end{abstract}

\section{Introduction}

Large Language Models (LLMs), such as GPT-4 and Llama 2, have demonstrated considerable potential in generating human-like text across a broad spectrum of fields, without additional domain-specific training. Their capabilities can be harnessed to provide assistance in the healthcare sector for a wide range of applications, including but not limited to disease diagnosis, clinical decision-making, and patient communication~\citep{cascella2023evaluating}. Despite the potential, the deployment of LLMs faces challenges. A prevalent concern is the tendency of LLMs to ‘hallucinate’, a term used to describe circumstances where the model generates plausible yet incorrect information, particularly when they are not able to provide an accurate response~\citep{Ji_2023}. In high-risk scenarios such as healthcare, where decisions can have direct impact on human lives, ensuring the reliability of LLMs becomes critical. This underscores the need for effective approaches to accurately estimate the uncertainty of generated responses and detect instances of hallucination.

In medical settings, existing methods for quantifying uncertainty, including entropy-based methods ~\citep{kadavath2022language, kuhn2023semantic} and fact-checking ~\citep{guo2022survey, shuster2021retrieval}, have demonstrated certain limitations. Entropy-based methods operate on the assumption that a model, when confident in its answer, generates a distribution of responses with a small entropy. On the contrary, if the model is unsure, it might hallucinate and produce a diverse range of responses, thus increasing the entropy ~\citep{kadavath2022language}. However, within the complexity of the medical domain, the model can often fabricate untruthful information with a high level of confidence. This results in a misleadingly low entropy, which fails to accurately represent the uncertainty embedded in the generated response. Fact-checking, another common approach for uncertainty estimation, validates the generated responses by comparing them with relevant truth retrieved from an external knowledge database. However, this method encounters limitations due to the scarcity of comprehensive and professional medical knowledge bases. 

\begin{figure}[ht]
    \centering
    
    \begin{subfigure}[a]{\textwidth}
        \includegraphics[width=\linewidth]{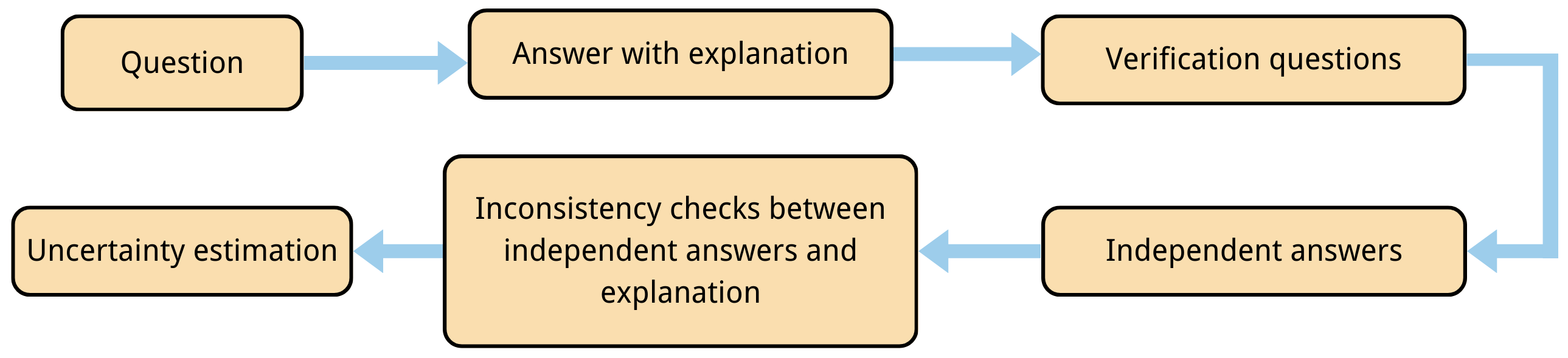}
        \caption{Chain-of-Verification (CoVe) method for Uncertainty Estimation}
        \label{cove-figure}
    \end{subfigure}
    \hfill 
    
    \begin{subfigure}[b]{\textwidth}
        \includegraphics[width=\linewidth]{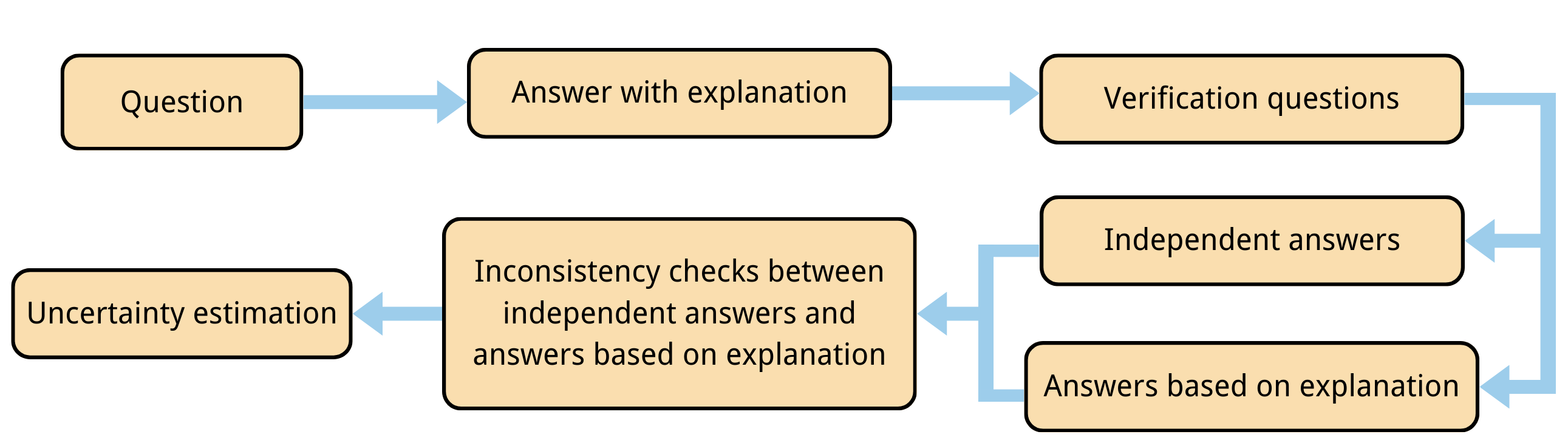}
        \caption{Two-phase Verification method for Uncertainty Estimation}
        \label{twophase-figure}
    \end{subfigure}
    
    \caption{Comparison of CoVe and Two-phase Verification Methods}
    \label{uncertainty_methods}
\end{figure}

In this report, we benchmark several popular methods using different model sizes and datasets to establish a comparative understanding of their performance. These benchmarks reveal the challenges of uncertainty estimation in medical question-answering. We also propose Two-phase Verification, a probability-free approach based on the \textit{Chain-of-Verification} (CoVe) concept ~\citep{dhuliawala2023chainofverification}. This approach operates independently of token-level probabilities and thus can be applied to black-box models. First, the model generates an explanation alongside its initial answer. Next, it formulates verification questions targeting the explanation, to which it provides independent answers. Two-phase Verification refines CoVe's inconsistency check process by prompting the model to answer the verification questions again, using the statement in question as a reference. The inconsistencies between the two sets of answers serve as a measure of uncertainty in the answer. The workflows of CoVe and Two-phase Verification are visualized in Figures \ref{cove-figure} and \ref{twophase-figure}, respectively.

\section{Related Work}

\subsection{Entropy-based methods}
Large Language Models generate output on a token-by-token basis based on the sequence so far. Token probabilities are a direct measure of a model's confidence in its next-token prediction. ~\citet{xiao2021hallucination} explore the link between hallucination in conditional language generation tasks and predictive uncertainty, which is quantified using entropy measures of the token probability distributions. They find that higher levels of predictive uncertainty, especially epistemic uncertainty originating from the model's knowledge gaps, correlate with an increased propensity for hallucinatory outputs. ~\citet{duan2023shifting} addresses the challenge of token-level generative inequality by a method termed Shifting Attention to Relevance (SAR), which reassigns attention to more semantically relevant components when estimating uncertainty. ~\citet{malinin2021uncertainty} introduce information-theoretic uncertainty measures at both the token and sequence levels and propose a novel metric, reverse mutual information, for structured uncertainty assessment, utilizing ensemble methods like Monte-Carlo Dropout and Deep Ensembles. ~\citet{kuhn2023semantic} propose semantic entropy, which measures uncertainty over meanings rather than just sequences of words. Their unsupervised method clusters semantically equivalent generations and calculates the predictive entropy of the resulting probability distribution over these clusters. A concurrent work of ~\citet{wang2024wordsequence} calibrates entropy-based uncertainty at word and sequence levels according to their semantic relevance, aiming to address the generative inequality challenge.

\subsection{Self-assessment methods}
LLMs possess the inherent potential to reflect on their outputs; however, the self-evaluation may not be robust as LLMs are inclined to find their own content credible~\citep{kadavath2022language}. To enhance the calibration and confidence estimation of LLMs, researchers have developed techniques including fine-tuning and prompting. ~\citet{lin2022teaching} finetune GPT-3 by supervised learning to express its uncertainty in natural language. Their experiment demonstrates that GPT-3 can be trained to provide answers along with a corresponding confidence level. Similarly, ~\citet{kadavath2022language} investigate the self-awareness of LLMs by training them to estimate the likelihood that their generated responses are correct. Their research reveals that the effectiveness of self-evaluation improves with model size and few-shot prompting. Additionally, the process benefits from presenting the model with various answer samples before asking it to assess the validity of a single proposed response. ~\citet{kojima2023large} explore the zero-shot capabilities of LLMs, finding that chain-of-thought prompting boosts the reasoning abilities of LLMs, especially in arithmetic tasks. By instructing LLMs to generate intermediate steps explicitly before answering the questions, a simple prompt template provides performance gain. ~\citet{manakul2023selfcheckgpt} introduce a zero-resource methodology for LLMs to self-check hallucinated generations based on the hypothesis that hallucinations tend to diverge. It operates by generating multiple responses to a prompt and then assessing the factual consistency between these responses.

\subsection{External tools}
Since knowledge gaps are a common cause of hallucinations, external knowledge retrieval is often utilized to mitigate hallucinations and produce more faithful generations ~\citep{he2022rethinking, shuster2021retrieval}. For example, ~\citet{chern2023factool} collect external evidence to validate the factuality of claims extracted from the LLM output. While prompting strategies enhance LLM performance in certain tasks, plausible explanations are often provided even when the final answer is wrong ~\citep{kojima2023large}. To overcome this limitation, ~\citet{chen2023chatcot} propose a multi-turn conversation framework to integrate prompting and external tools including calculators and search engines, which reduces the mistakes made by LLMs and enhances the accuracy in complex reasoning tasks.

\section{Methodology}
In this section, we elaborate on our approach to estimating generation uncertainty, which leverages the idea from the Chain-of-Verification (CoVe) framework. Our approach is inspired by the foundational work of ~\citet{dhuliawala2023chainofverification} and extends it by integrating a measure for confidence level based on discovered inconsistencies. The primary goal is to identify the occurrence of possible hallucinations by incorporating a robust, unsupervised verification mechanism that operates independently of the model's initial outputs.

\subsection{Generate step-by-step explanation}
For each question, the LLM is required to generate a definitive answer followed by a step-by-step explanation. We perform the experiment on two types of questions: those that require a ternary response (affirmative, negative, or uncertain) and those that present multiple-choice options. The definitive answer will be in the form of "yes," "maybe," or "no" for the first type of questions, or a selection from the multiple-choice options for the second type. This is followed by generating a detailed step-by-step explanation for the chosen answer, which is critical for the subsequent verification chain. The step-by-step breakdown converts the model's reasoning into discrete units that can be independently verified for truthfulness and consistency, thereby enabling an estimation of the overall confidence in the response.

\subsection{Plan verification}
Upon generating the initial answer and step-by-step explanation, the model proceeds to formulate a set of verification questions, with each one targeting a single step in the explanation. These questions are purposefully designed to challenge the accuracy of particular factual claims within the individual steps of the explanation. The objective of these questions is to verify the truthfulness of each assertion without necessitating supplementary knowledge or additional context for their resolution. For example, in response to the statement \textit{"Ringed sideroblasts are a characteristic feature of iron overload, particularly in the bone marrow"}, a potential verification question could be \textit{“What condition are ringed sideroblasts typically indicative of?”} This question directly targets the factual claim made within the statement and is structured to elicit a response that either confirms or refutes the accuracy of the original statement.

While the model is capable of formulating reasonable verification questions on a zero-shot instruction, the incorporation of a few-shot prompt significantly refines this procedure by enhancing the efficacy of the verification questions. A few-shot prompt presents the model with a set of carefully curated exemplary pairs of statements and corresponding verification questions. These examples serve as a template, showcasing the structure and purpose of a well-crafted verification question. Consequently, this few-shot prompt empowers the model to formulate questions that are not only relevant but also incisive in their ability to discern and test the validity of factual assertions.

\subsection{Execute verification}
Given the verification questions, in the next step, the model executes the verification procedure to self-check whether the explanation is accurate. We examined several different approaches for verification in our experiment.

\subsubsection{Step verification} As a base for the verification procedure, we use the model to directly assess the truthfulness and the consistency related to the previous steps of each sentence in the explanation without utilizing the verification questions. For each sentence, the model is prompted to determine its truthfulness based on the prior sentences in the explanation, classifying it as true or false. This serves as a baseline measurement of the model's ability to self-validate its content.

This direct approach assumes that the language model is intrinsically capable of recognizing factual information. It provides a straightforward validation mechanism without the additional layer of complexity introduced by verification questions.

\subsubsection{CoVe} In this approach, the LLM answers the verification questions independently to avoid the influence of the initial output. Next, the independent answer will be checked against the original statement being examined for consistency. This is performed by providing the model with both the answer and the statement and asking it to decide if they are consistent or not.

The assumption for this approach is that the model is less likely to repeat any hallucinations present in the initial explanation when answering the verification questions independently without any context. If the independent response aligns with the explanation, the corresponding statement has a lower possibility of being a hallucination. On the contrary, an inconsistency between the two indicates a potential error or hallucination in the explanation, making the initial answer less plausible.

\subsubsection{Two-phase verification} 
\begin{figure}[h]
\begin{center}
\includegraphics[width=\textwidth]{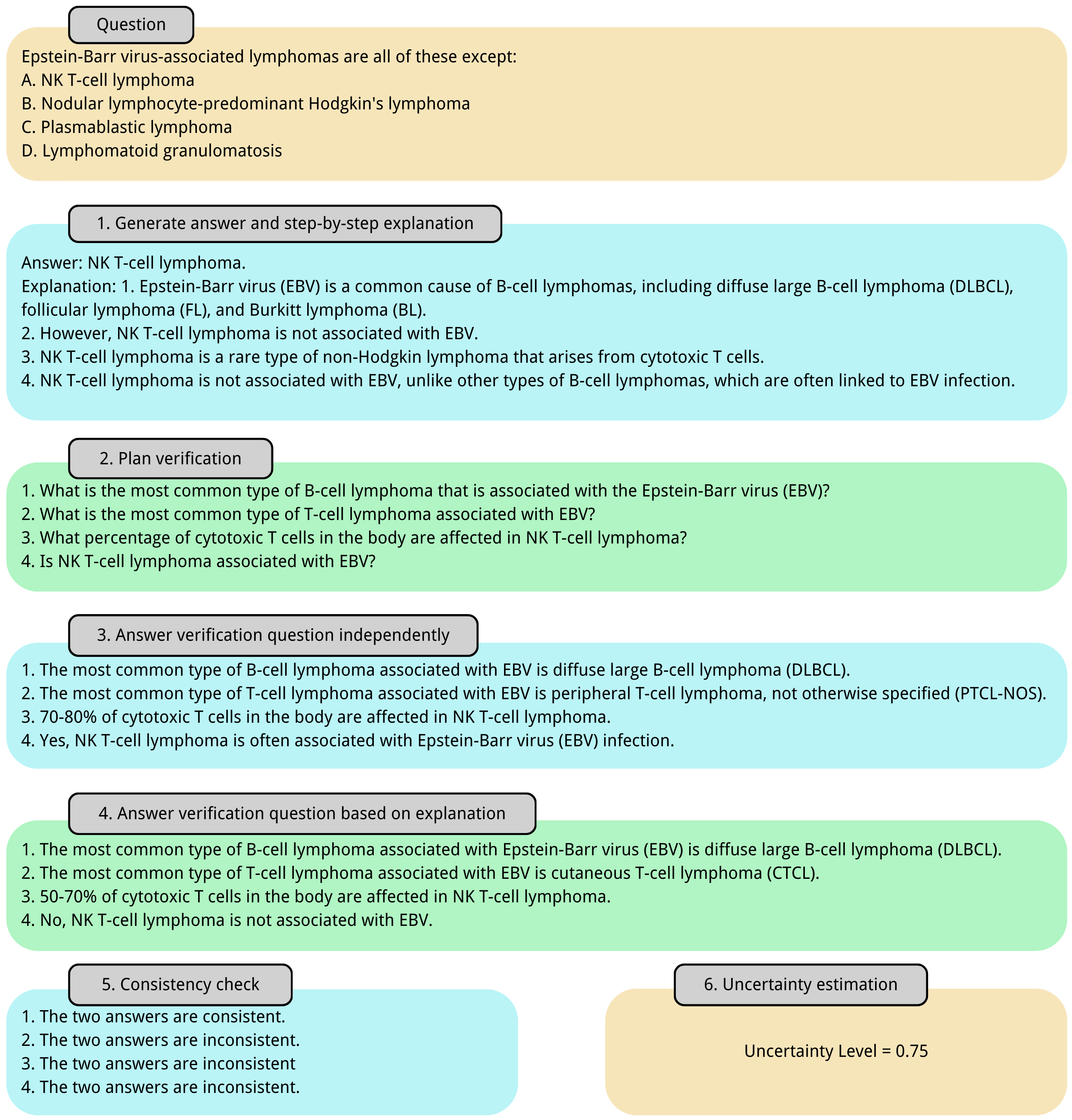}
\end{center}
\caption{Illustration of Two-phase Verification process with an example question}
\label{process-figure}
\end{figure}

In this more sophisticated approach, the model is prompted to answer each verification question twice. First, the model answers the verification question independently, as in the previous approach. Next, the model is given the statement to be verified as the context and prompted to answer the verification question again. To evaluate whether the two answers are consistent, we adopt a method for checking semantic equivalency which uses a Deberta-large model \citep{he2021deberta} for a bidirectional entailment check \citep{kuhn2023semantic}. This process involves appending a special token between the answers and evaluating whether each answer can be inferred from the other, with equivalence determined by mutual "entailment" classifications by the model. An example of the Two-phase Verification procedure is illustrated in Figure \ref{process-figure}.

The rationale for integrating the second verification question answering step responds to two significant challenges encountered during the consistency check in CoVe:
\begin{enumerate}
\item Ambiguity in Consistency Checks:
The instructions for a consistency check can themselves be ambiguous due to the different interpretations of “consistency”. Additionally, the model may fixate on superficial linguistic patterns rather than the underlying factual content. Various phrasings conveying the same meaning may not be recognized as consistent by the model, leading to false judgments in identifying consistency.

\item Relevance and Information Discrepancies:
The independent answer generated by the model could introduce additional information that is not strictly relevant to the initial explanation, or it could omit crucial details, making it difficult to accurately assess consistency. The answer might be factually correct in itself but still not align perfectly with the explanation due to differences in scope or detail level.

\end{enumerate}

\subsection{Uncertainty quantification}
Following the completion of the verification steps, we translate the findings into a measurable indicator of uncertainty. This involves counting the statements identified as inconsistent in the verification phase, relative to the overall number of statements in the provided explanation. To express this quantitatively, we compute the Uncertainty Level (UL) using the formula below: $$ UL=\frac{\text{Number of Inconsistent Statements}}{\text{Total Number of Statements in Explanation}} $$

\section{Experiment}
\subsection{Experimental setting}
\textbf{Models} We conduct the experiment on Llama 2 Chat \citep{touvron2023Llama}, which is a collection of open-source chat models fine-tuned for optimized dialogue use cases. Llama 2 Chat (7b) and Llama 2 Chat (13b) were examined in our experiment.

\textbf{Datasets} We consider three biomedical question-answering (QA) datasets: PubMedQA \citep{jin-etal-2019-pubmedqa}, MedQA \citep{jin2021disease} and MedMCQA \citep{pmlr-v174-pal22a}. \textbf{PubMedQA} is a biomedical research QA dataset designed to answer questions with a yes/no/maybe format. Each question comes with a context extracted from the corresponding abstract of a research paper and challenges models to reason over quantitative biomedical content. The expert-annotated questions were utilized in our experiment. \textbf{MedQA} is a free-form multiple-choice QA dataset collected from questions in professional medical board examinations, such as the United States Medical Licensing Examination (USMLE). \textbf{MedMCQA} is a large-scale multiple-choice QA dataset derived from real-world medical entrance exam questions. It is designed to test a variety of reasoning abilities across a wide range of medical subjects and topics. 

\textbf{Baselines} We consider 4 baseline methods in our experiments, including  Lexical Similarity (LS) \citep{fomicheva-etal-2020-unsupervised}, Semantic Entropy (SE) \citep{kuhn2023semantic}, Predictive Entropy (PE) \citep{kadavath2022language} and Length-normalized Entropy (LE) \citep{malinin2021uncertainty}. \textbf{Lexical Similarity} among a set of generated texts is quantified by computing the average ROUGE-L score, and a higher similarity score indicates that the model is more certain in its responses. \textbf{Semantic Entropy} addresses the difficulty of semantic equivalence in the uncertainty estimation of free-form LLMs by clustering generations with the same semantic meanings and calculating cluster-wise entropy.  \textbf{Predictive Entropy} is estimated by averaging the sum of negative log probabilities of each token in the sampled answers for a given question. \textbf{Length-normalized Entropy} divides the sum of negative log probabilities of a sequence by its length, handling the issue of disproportionate contribution to the total entropy due to variable sentence length. 

\textbf{Metrics} Following \citet{kuhn2023semantic}, we evaluate the performance of our uncertainty estimation approach using the area under the receiver operating characteristic curve (AUROC) as our metric. The metric measures the probability that a randomly chosen correct answer has a lower uncertainty level compared to a randomly chosen incorrect answer. 

\subsection{Results}
\begin{table}[t]
\caption{AUROC results for various uncertainty estimation methods across multiple datasets and model sizes. Methods include Lexical Similarity (LS), Semantic Entropy (SE), Predictive Entropy (PE), Length-normalized Entropy (LE), Step Verification (Step), Chain-of-Verification (CoVe) and Two-phase Verification (Two-phase). Results are shown for two model sizes: Llama 2 Chat (7b) and Llama 2 Chat (13b), evaluated on PubMedQA, MedQA, and MedMCQA datasets. The highest AUROC score for each model-dataset combination and the overall best results for averages and standard deviations (SDs) across datasets are highlighted in bold. For entropy-based methods, 5 answers are generated for each question, and the temperature is set to 0.5, which optimized Semantic Entropy (SE) and Length-normalized Entropy (LE) \citep{kuhn2023semantic}.}
\begin{center}
\begin{tabular}{lccccccc}
\toprule
\multicolumn{1}{l}{}  &\multicolumn{1}{c}{LS\footnotemark}  &\multicolumn{1}{c}{SE\footnotemark} &\multicolumn{1}{c}{PE\footnotemark} &\multicolumn{1}{c}{LE\footnotemark} &\multicolumn{1}{c}{Step} &\multicolumn{1}{c}{CoVe\footnotemark} &\multicolumn{1}{c}{Two-phase (Ours)}  \\
\midrule
\midrule
\multicolumn{8}{l}{\textbf{Llama 2 Chat (7b)}} \\
\midrule
PubMedQA & 0.5277 & 0.6320 & 0.6322 & 0.6028 & 0.6288 & \textbf{0.6866} & 0.6132 \\
MedQA & 0.4871 & 0.5154 & 0.5189 & 0.5224 & 0.4170 & 0.4861 & \textbf{0.5553} \\
MedMCQA & 0.3837 & 0.4676 & 0.5028 & \textbf{0.6013} & 0.5178 & 0.5509 & 0.5304 \\

\midrule
\textbf{Average} & 0.4662 & 0.5383 & 0.5513 & \textbf{0.5755} & 0.5212 & 0.5745 & 0.5663 \\
\textbf{SD} & 0.0742 & 0.0846 & 0.0705 & 0.0460 & 0.1059 & 0.1023 & \textbf{0.0425} \\
\midrule
\midrule

\multicolumn{8}{l}{\textbf{Llama 2 Chat (13b)}} \\
\midrule
PubMedQA & 0.5551 & 0.5689 & 0.5681 & 0.4503 & 0.5085 & 0.5352 & \textbf{0.5906} \\
MedQA & 0.4860 & 0.4898 & 0.4010 & 0.5077 & 0.5934 & 0.5408 & \textbf{0.6460} \\
MedMCQA & 0.5142 & 0.5247 & 0.5708 & 0.5933 & 0.4895 & \textbf{0.6026} & 0.5793 \\
\midrule
\textbf{Average} & 0.5184 & 0.5278 & 0.5133 & 0.5171 & 0.5305 & 0.5595 & \textbf{0.6053} \\
\textbf{SD} & \textbf{0.0347} & 0.0396 & 0.0973 & 0.0720 & 0.0553 & 0.0374 & 0.0357 \\
\midrule
\midrule
\textbf{Overall average} & 0.4923 & 0.5331 & 0.5323 & 0.5463 & 0.5258 & 0.5670 & \textbf{0.5858} \\
\textbf{Overall SD} & 0.0592 & 0.0593 & 0.0788 & 0.0628 & 0.0758 & 0.0694 & \textbf{0.0411} \\
\bottomrule
\end{tabular}

\end{center}

\label{result-table}
\end{table}
\footnotetext[1]{Lexical Similarity \citep{fomicheva-etal-2020-unsupervised}}
\footnotetext[2]{Semantic Entropy \citep{kuhn2023semantic}}
\footnotetext[3]{Predictive Entropy \citep{kadavath2022language}}
\footnotetext[4]{Length-normalized Entropy \citep{malinin2021uncertainty}}
\footnotetext[5]{Chain-of-Verification \citep{dhuliawala2023chainofverification}}

\begin{figure}[t]
\begin{center}
\includegraphics[width=0.5\textwidth]{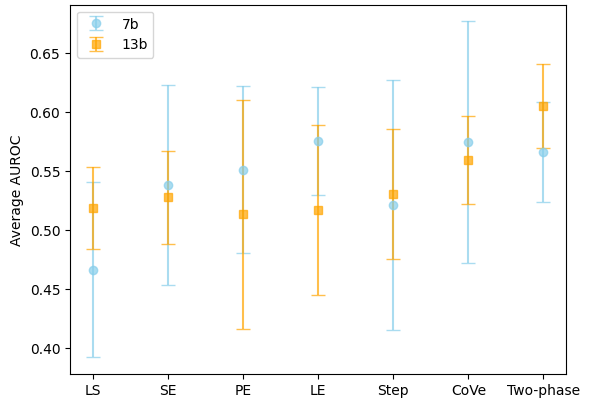}
\end{center}
\caption{Performance comparison of UE methods on different model sizes}
\label{performance-figure}
\end{figure}
We compare Two-phase Verification with several baseline methods on three medical datasets using two Llama 2 Chat models. The results are summarized in Table \ref{result-table} and Figure \ref{performance-figure}.

Lexical Similarity (LS), which assesses uncertainty based on the overlap among sample responses, shows the lowest overall average AUROC. This suggests that lexical resemblances are insufficient indicators of certainty in the generated text, where semantic meaning is crucial. Semantic Entropy (SE) and Predictive Entropy (PE) demonstrate moderate improvements over LS. The two methods achieve similar AUROC scores, as they both estimate uncertainty from the entropy of sample responses. SE has slightly better overall results than PE, indicating that semantic clustering is an effective strategy in entropy-based methods. Length-normalized Entropy (LE) achieves the highest average AUROC for the Llama 2 Chat (7b) model, suggesting that normalizing entropy by answer length provides a more reliable uncertainty signal for smaller models. However, LE's performance does not consistently hold across larger model sizes or all datasets, indicating that while length normalization is beneficial, it is not a comprehensive solution for UE.

Step Verification, as the most straightforward verification method that relies solely on the model's self-validation ability, does not show a performance improvement compared to other baseline methods. Interestingly, for the 7b model, the performance of Step Verification appears to be correlated with the accuracy of the model's answers on each dataset. PubMedQA, which has the highest answer accuracy (0.65), also shows the best Step Verification result, while MedQA and MedMCQA, with lower accuracies (0.2991 and 0.3429, respectively), have poorer Step Verification performance. This observation suggests that smaller models may exhibit overconfidence in their generated answers and struggle to identify their own mistakes during self-verification. This limitation highlights the necessity of introducing a verification chain to help the model recognize hallucinations in its outputs.

CoVe slightly outperforms Two-phase Verification in some cases but has a high standard deviation, particularly with smaller model sizes. This may stem from the variability in the quality of independently generated answers, as smaller models might produce less faithful responses. Further research could explore ways to improve the reliability of answering verification questions, potentially by integrating external knowledge bases.

In general, Two-phase Verification demonstrates the best overall performance, achieving the highest AUROC in half of the model-dataset combinations and the highest average AUROC across all experiments. It also exhibits stable performance with the lowest overall standard deviation, unlike other methods such as CoVe, which show performance fluctuations in certain scenarios. Moreover, the scalability of Two-phase Verification is particularly noteworthy when comparing the AUROC results for Llama 2 Chat (7b) and Llama 2 Chat (13b). While most methods show only modest improvements or, in some cases, a decrease in performance with the larger model size, Two-phase Verification not only improves but also does so at a higher rate than its counterparts. These characteristics suggest the method's potential to provide reliable uncertainty estimation across various datasets and to scale with larger model sizes.

\section{Discussion}
\subsection{Uncertainty Estimation in medical QA}
Uncertainty Estimation is of paramount importance in the medical domain, where untruthful information can lead to severe consequences. In medical applications utilizing LLMs, such as AI medical chatbots, it is crucial to assess the trustworthiness of model outputs to ensure patient safety. In the cases where the model is less certain in its predictions, the user should be alerted and advised to seek further verification or expert opinion before following the model's suggestions.

The findings of our empirical study contribute to the literature on UE of LLMs, particularly in the context of medical question answering, which has been less studied. Previous research has primarily focused on UE from a statistical perspective, hypothesizing that the model intrinsically knows when it is uncertain about an answer, leading to higher variability in its outputs. However, professional medical knowledge is often underrepresented in the training data, which can lead to the model generating responses confidently even when it is hallucinating. As a result, answers may exhibit low entropy, falsely suggesting high certainty.

Our Two-phase Verification method mitigates these issues by independently verifying responses, thus providing an effective measure of a model's certainty without needing token-level probabilities. This is especially useful for black-box models, where architectural details are inaccessible. Moreover, the scalability of the Two-phase Verification method is a critical aspect for future applications. As large-scale models continue to evolve, the ability to maintain and even enhance performance with increased model size is essential.

The concept of \textit{Chain-of-Verification} (CoVe) has been previously proposed to reduce hallucinations and then self-correct them to generate more factual statements \citep{dhuliawala2023chainofverification}. However, to the best of our knowledge, this concept has not been explored for uncertainty estimation. Our work demonstrates the effectiveness of integrating a verification chain for uncertainty estimation in medical question-answering, opening up new possibilities for future research in this direction.

\subsection{Limitations and future work}
\textbf{Verification question generation} A critical stage of Two-phase Verification is to generate verification questions that effectively challenge the initial explanation. As explanation paragraphs are generated with linguistic coherence, sentences often use pronouns or references that rely on previous sentences. When verification questions are derived from discrete sentences, the model may miss essential context. Thus, the verification questions might not always incisively interrogate the key information presented. Although few-shot prompts aid question formulation, they can inadvertently inhibit the LLM's creativity, confining it to the patterns seen in these examples. In future work, it will be essential to enhance the generation of verification questions to be more context-aware and adaptable.

\textbf{Domain knowledge constraints} Another constraint for Two-phase Verification is the knowledge capacity of the language model, which directly affects the quality of the answers to verification questions. Llama 2 Chat, as a general-purpose language model, possesses only a broad understanding of medical knowledge, lacking the depth required for specialized areas. To improve the model's responses to verification questions, we integrate dense retrieval techniques to source relevant information from external databases like Wikipedia. However, this method falls short as the retrieved results frequently have low relevance scores to the verification queries and fail to provide the necessary knowledge. Future improvements should focus on retrieving relevant information from professional medical datasets, such as research papers, medical textbooks, and expert-curated knowledge bases. By leveraging these domain-specific resources, the model can generate more accurate and reliable independent answers to verification questions, enabling more effective detection of hallucinations and uncertainties in medical explanations.

\section{Conclusion}
In this paper, we conduct an empirical study on the Uncertainty Estimation of LLMs in medical question-answering tasks. We find Uncertainty Estimation challenging in the medical domain, with existing methods performing poorly, especially with smaller model sizes. To address this challenge, we propose Two-phase Verification, a novel approach that integrates the concept of CoVe to assess the reliability of language model outputs. We show that the model is capable of detecting its own hallucinations by answering verification questions independently and cross-checking against the answers referencing its initial reasoning. Overall, our Two-phase Verification method demonstrates superior performance over baseline methods and is reliable across various model sizes and dataset settings.

\bibliography{colm2024_conference}
\bibliographystyle{colm2024_conference}


\end{document}